\documentclass{article}
\usepackage{spconf,amsmath,graphicx}
\usepackage{booktabs}
\usepackage[utf8]{inputenc}
\usepackage[linesnumbered,ruled,vlined]{algorithm2e}
\usepackage{url}
\usepackage{setspace} 
\usepackage{caption}
\usepackage[numbers]{natbib}
\usepackage{float}
\usepackage{fancyhdr}
\title{Improve Deep Forest with Learnable Layerwise \\
Augmentation Policy Schedules}
\name{Hongyu Zhu\textsuperscript{1}, Sichu Liang\textsuperscript{2}, Wentao Hu\textsuperscript{1}, Fang-Qi Li\textsuperscript{3}, Yali yuan\textsuperscript{1}, Shi-Lin Wang\textsuperscript{3}, Guang Cheng\textsuperscript{1}}

\address{\textsuperscript{1} School of Cyber Science and Engineering,
\textsuperscript{2} School of Artificial Intelligence,
Southeast University.\\
\textsuperscript{3} School of Electronic Information and Electrical Engineering,
Shanghai Jiao Tong University.}
\begin{document}
\fancypagestyle{copyrightpage}{%
  \fancyfoot[C]{\fontsize{9}{9}\selectfont \textcopyright{} 2023 IEEE. Personal use of this material is permitted. Permission from IEEE must be obtained for all other uses, in any current or future media, including reprinting/republishing this material for advertising or promotional purposes, creating new collective works, for resale or redistribution to servers or lists, or reuse of any copyrighted component of this work in other works.}%
}
\thispagestyle{copyrightpage}
%
\maketitle
\begin{abstract}
As a modern ensemble technique, Deep Forest (DF) employs a cascading structure to construct deep models, providing stronger representational power compared to traditional decision forests. However, its greedy multi-layer learning procedure is prone to overfitting, limiting model effectiveness and generalizability. This paper presents an optimized Deep Forest, featuring learnable, layerwise data augmentation policy schedules. Specifically, We introduce the Cut Mix for Tabular data (CMT) augmentation technique to mitigate overfitting and develop a population-based search algorithm to tailor augmentation intensity for each layer. Additionally, we propose to incorporate outputs from intermediate layers into a checkpoint ensemble for more stable performance. Experimental results show that our method sets new state-of-the-art (SOTA) benchmarks in various tabular classification tasks, outperforming shallow tree ensembles, deep forests, deep neural network, and AutoML competitors. The learned policies also transfer effectively to Deep Forest variants, underscoring its potential for enhancing non-differentiable deep learning modules in tabular signal processing.
\end{abstract}
%
\vspace{-1.0em}
\section{INTRODUCTION}
\vspace{-0.8em}
\label{sec:intro}
In recent years, deep neural networks (DNNs) have achieved remarkable success in perception tasks such as vision, speech, and language \cite{lecun2015deep}. However, heterogeneous tabular data in industrial scenarios still poses significant challenges, making it the "last unconquered castle" for DNNs \cite{kadra2021well}. In tabular modeling tasks, shallow decision tree ensembles like random forests and gradient boosting trees remain the preferred tools for data mining specialists \cite{NEURIPS2022_0378c769}. However, these methods lack the ability to learn deep representations, which limits their potential to benefit from larger models and more training data \cite{feng2018multi,Popov2020Neural, 10096564}.

Drawing on the core principles of deep learning—layer-by-layer processing, intra-model feature transformations, and sufficient capacity—Zhou et al. introduced Deep Forest (DF), a tree-based deep learning approach \cite{zhou2017deep}. Deep Forest constructs a cascading structure by training random forests layer by layer. In classification tasks, each layer's forests generate class probability distributions, which are then merged with original features for the subsequent layer. Upon reaching the maximum layer depth, optimal layer selected by cross-validation is used to produce final output. When sufficiently
deep, Deep Forest aims to achieve robust and powerful feature representations through this iterative refinement \cite{song2021sparse,zhang2019deep,wen2019multi}.

However, due to its supervised greedy multi-layer architecture, Deep Forest is prone to overfitting \cite{lyu2022depth}, and such issue in
one lay can propagate to subsequent layers \cite{arnould2021analyzing}. While increased depth should theoretically enhance representational power \cite{lyu2022depth}, it frequently leads to severe overfitting in practice, undermining generalizability \cite{chen2021improving}. Moreover, the similar learning objectives across layers diminish the benefits of building deeper models.

Overfitting in Deep Forest stems from its intrinsic learning procedure, which is difficult to suppress through hyperparameter tuning \cite{arnould2021analyzing}. Reducing the model size, on the other hand, would compromise its design principle. Data augmentation offers an alternative for regularization without limiting model capacity \cite{geiping2023how}. However, tabular data augmentation remains underexplored due to the absence of invariances in heterogeneous features \cite{gordon2022data}. Furthermore, uniform augmentation intensity across layers result in similar feature representations and introduce high variance, complicating selection of the optimal layer based on validation error.

In this work, we propose to improve Deep Forest with learnable data augmentation policy schedules. Firstly, we introduce a simple yet potent data augmentation technique called CMT (\textbf{C}ut \textbf{M}ix for \textbf{T}abular data) to regularize Deep Forest. Furthermore, a population-based algorithm is proposed to globally search for augmentation policy schedules, enabling layer-specific adjustments in augmentation intensity. Finally, we utilize outputs from intermediate layers to construct a checkpoint ensemble, serving as a variance reducer to ensure more stable and robust results.

Our contributions are summarized as follows:
{
\vspace{-0.8em}
\begin{itemize}
    \setlength{\itemsep}{-5pt}
    \setlength{\parsep}{0pt}
    \setlength{\topsep}{0pt}
    \item We introduce tabular data augmentation, specifically CMT, to mitigate overfitting in Deep Forest.
    \item We optimize augmentation policy schedules through population-based algorithm, with moderate overhead.
    \item We improve Deep Forest by integrating outputs from each layer to form a stable checkpoint ensemble.
    \item Our method sets SOTA benchmarks in tabular classification and enables policy transfer to DF variants.
\end{itemize}
}

{
\setlength{\intextsep}{-10pt} 
\setlength{\abovecaptionskip}{-10pt}
\begin{figure}
    \centering
    \includegraphics[width=\linewidth]{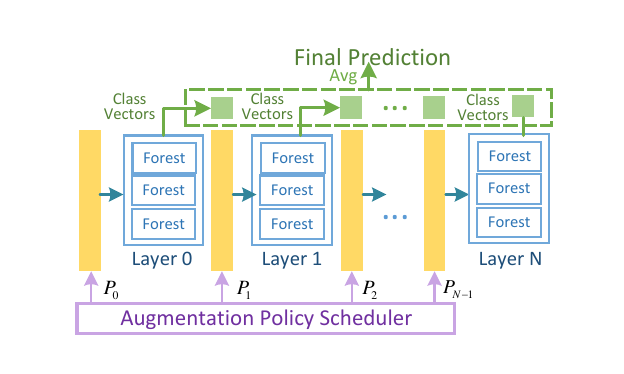}
    \caption{AugDF architecture with learned policy schedule.}
    \label{fig:stru}
    \vspace{-15pt} 
\end{figure}

}
{
\vspace{-15pt}
\section{METHODOLOGY}
\vspace{-5pt}
}
\subsection{Cut Mix for Tabular Data (CMT)}
\vspace{-0.5em}

Data Augmentation, guided by the Vicinal Risk Minimization principle \cite{NIPS2000_ba9a56ce}, has significantly advanced regularization of DNNs in fields like vision and language \cite{geiping2023how}. However, the tabular domain still lacks effective augmentation techniques \cite{gordon2022data}. The mixup method \cite{DBLP:conf/iclr/ZhangCDL18}, widely used across data modalities, generate new samples through linear combination of original instances. However, mix-up has two primary limitations in this context. First, convex combination of categorical variables can lead to semantic ambiguity. Second, tabular data exhibits irregular patterns \cite{NEURIPS2022_0378c769}, and samples constructed through simple linear interpolation may introduce bias.

We thus design Cut Mix for Tabular data (CMT) to address these issues. The new samples are formulated as:
{
\vspace{-7pt}
\begin{equation}
\setlength{\intextsep}{-10pt}
\vspace{-7pt}
\begin{aligned}
\tilde{x} &= \mathbf{w} \odot x_i + (\mathbf{1} - \mathbf{w}) \odot x_j \\
\tilde{y} &= c \cdot y_i + (1 - c) \cdot y_j
\end{aligned}
\end{equation}

}

Here, $x_i,x_j\in\mathbf{R}^d$ are original samples, and $y_{i}$,$y_{j}$ are their one-hot encoded labels. $\mathbf{w}\in{\left\{ {0,1} \right\}^d}$ is a random mask vector with $\|\mathbf{w} \|_1 = \lambda \cdot d$, $\lambda \sim \text{Beta}(\alpha, \alpha)$. The coefficient $c$ 
is calculated by the Feature Importance (FI) vector derived from decision forests in the preceding layer, where $FI(k)$ indicates importance of the $kth$ feature:
\vspace{-8pt}
\begin{equation}
\vspace{-5pt}
c = \frac{\sum_{k=1}^{d} w_k \cdot FI(k)}{\sum_{k=1}^{d} FI(k)}
\label{equation}
\end{equation}
The blending strategy of CMT swaps partial features between samples, drawing all values from the original dataset to avoid semantic ambiguity in categorical variables. Moreover, this method extends beyond linear combinations, better accommodating the diverse patterns in heterogeneous tabular data.
\vspace{-2.5em}
\subsection{Augmentation Policy Schedule Learning}
\label{APSL}
\vspace{-0.5em}

The intensity of data augmentation is a pivotal factor affecting model performance, as excessive or inappropriate augmentation can introduce biases \cite{cubuk2019autoaugment,ho2019population}. To regulate this, an augmentation policy \(\theta\) is optimized as follows:
{
\vspace{-7pt}
\begin{equation}
\theta^{*}=\underset{\theta \in \Theta}{\arg \max } \operatorname{eval}(\theta)
\label{equation}
\end{equation}
\vspace{-7pt}
}

where \(\theta\) consists of hyperparameters \(prob\) and \(mag\), representing the likelihood of sample selection for augmentation and the perturbation magnitude for each selected sample ($\alpha$ in CMT).
\(prob\) and \(mag\) are respectively taken from lists P = $\left[ {{p_1},{p_2} \ldots {p_M}} \right]$ and M = $\left[ {{m_1},{m_2} \ldots {m_N}} \right]$, resulting in a searching space of size \(M \times N\).  \(\operatorname{eval}(\theta)\) represents the objective to optimize, i.e. accuracy score on the validation set.


While a static policy provides certain regularization benefits, it is limited in mitigating inter-layer convergence. We thus reformulate the problem to identify an optimal sequence of policy combinations, establishing a layerwise policy schedule for Deep Forest. The policy for the \(k\)-th layer is denoted by \(\theta^k = ({prob}^k, {mag}^k)\). In a DF with \(K\) layers, the complete policy schedule is represented as \(\Theta = (\theta^1, \theta^2, \ldots, \theta^K)\). This approach enlarges the search space to \((M \times N)^K\), offering much greater flexibility for augmentation.

However, navigating through such a vast search space is computationally intractable. Therefore, we propose a population-based search algorithm optimized for the unique attributes of DF while ensuring low computational overhead. Initially, we formulate two single-layer search strategies:

\textbf{Grid Search}: This method scans all hyperparameter combinations to maximize current layer's validation accuracy, but is feasible only for a single layer due to complexity.

\textbf{Neighbour Search}: 
Given a selected policy ${\theta _{i,j}} = \left( {{p_i},{m_j}} \right)$  from the set $P \times M$, a neighboring policy ${\theta _{x,y}} = \left( {{p_x},{m_y}} \right)$ is chosen with probability of \(\frac{1}{|i-x| + |j-y| + 1}\). After normalizing these probabilities, multiple sampling rounds are performed to obtain the required number of different policies.

In the population, $2k$ DFs are considered, each acting as an individual. We initialize the best policy \(\theta^0\) for the first layer with grid search and acquire $2k-1$ neighboring policies via neighbour search. These $2k$ policies are used to train the first layer of each DF. For subsequent layers, the procedure iteratively continues as follows:

\textbf{Explore}: Leveraging the policies of the current half top-performing individuals, an exploration by neighbour search is conducted to generate a set of $k$ new policies. These new policies replace the existing $k$ 
inferior ones, and each individual is then trained with this updated policy set.

\textbf{Exploit}: The current validation accuracy of each individual is assessed and ranked. The upper half of top-performing individuals, along with their associated policies, are preserved, while the lower-performing half is substituted by the current best individual with explored policies.


Upon iterating through each layer, a complete policy schedule is finalized. See Algorithm 1 for pseudocode.
\vspace{-10pt}
\subsection{Checkpoint Ensemble (CE)}
\vspace{-5pt}
{
\begin{algorithm}[h]
\caption{The process of Augmentation Policy Schedule Learning. Functions grid search and neighbour search are detailed in section \ref{APSL}. Explore and exploit phases are outlined in the same section.}
\SetKwInOut{Input}{Input}
\SetKwInOut{Output}{Output}
\SetKwFunction{GridSearch}{grid\_search}
\SetKwFunction{NearbySearch}{neigh\_search}
\SetKwFunction{Eval}{eval}
\SetKwProg{Function}{Function}{:}{}
\Input{list of policies $p$, list of models $m$, number of models (policies) $L$, max layer $K$}
\Output{Optimal model $m^*$}

$p[0] \leftarrow \GridSearch{}$\tcp*[r]{initialize}
$p[1:L-1] \leftarrow \NearbySearch{p[0],L - 1}$\;
$m, p \leftarrow \Eval{m,p}$\;
\For{ $i=1$ \KwTo $K-1$}{
    $m[L/2:L-1] \leftarrow [m[0]]* L / 2$\tcp*[r]{exploit}
    $p[L/2:L-1] \leftarrow \NearbySearch{p[0],L / 2}$\;
    $m, p \leftarrow \Eval{m,p}$\tcp*[r]{explore}
}

\KwRet $m[0]$\;

\Function{\GridSearch{}}{
    find the best policy $bp$ using grid search\;
    \KwResult{bp}
    \BlankLine
}

\Function{\NearbySearch{bp, X}}{
    generate $X$ new policies $[new\_policies]$ nearby current best policy $bp$\;
    \KwResult{[new\_policies]}
    \BlankLine
}
\Function{\Eval{m, p}}{
    Train each $model$ in $m$ corresponding to $p$, evaluate the accuracy of each $model$, and sort $m$ and $p$ by accuracy in descending order\;
    \KwResult{m,p}
}
\end{algorithm}
}
The concept of using "historical" information in deep learning is well-established \cite{DBLP:journals/corr/abs-2303-12992}. Averaging snapshots from the SGD trajectory often improves generalization \cite{huang2017snapshot,garipov2018loss,izmailov2018averaging}. While snapshot ensemble enhances performance, it increases storage and inference latency due to multiple weight sets and 
evaluations. In AugDF, we harness the hierarchical diversity introduced by the augmentation policy schedule to construct a checkpoint ensemble, as illustrated in Figure \ref{fig:stru}, which incurs neither additional training cost nor significant inference overhead. This strategy serves as a variance reducer, while further boosting the representational capability of the model, thereby achieving ensemble benefits (almost) for free.

The processing procedure of a test sample $x$ in Deep Forest can be recursively defined as follows:
{
\setlength{\intextsep}{-5pt}
\vspace{-5pt}
\begin{equation}
\vspace{-5pt}
{F_i}\left( x \right) = \left\{ \begin{array}{l}
{h_i}\left( x \right),\;\;i = 1\\
{h_i}\left( {x;{F_{i - 1}}\left( x \right)} \right),\;\;1 < i \le K
\end{array} \right.
\label{equation}
\end{equation}
}
where ${h_i}$ represents the $i-th$ layer of the Deep Forest, and ${F_i}$ denotes the model up to the $i-th$ layer. $K$ is the maximum number of layers in DF. Unlike the conventional approach where the final output layer $o$ is determined through cross-validation, resulting in ${F_o}\left( x \right)$ as the output, the checkpoint ensemble can be represented as:
{
\setlength{\intextsep}{-5pt}
\vspace{-5pt}
\begin{equation}
\vspace{-5pt}
{F_{CE}} = \frac{1}{K}\sum\nolimits_{i = 0}^{K - 1} {{F_{K - i}}\left( x \right)}
\label{equation}
\end{equation}
}

    Note that intermediate layers in Deep Forest already require computation; thus, adding a checkpoint ensemble incurs no extra training overhead. While early-stopping is possible through cross-validation, deeper models can be more powerful \cite{lyu2022depth}. Therefore, DF in practice often approach maximum depth \cite{lyu2019refined}. Accordingly, the inference overhead of a checkpoint ensemble is essentially the same as a vanilla DF.

{
\vspace{-1.2em}
\section{EXPERIMENTS}
\vspace{-0.8em}
}
\textbf{Dataset. }
Experiments are conducted on eight benchmark datasets for binary and multi-class classification tasks from the UCI Machine Learning Repository \cite{Dua:2019}. These datasets vary significantly in size, ranging from 452 to 1,516,948 instances, and span multiple application domains such as ECG signals for cardiac arrhythmia diagnosis, remote sensing signals for covertype recognition, air production unit signals for failure prediction, and mobile phone signals for activity recognition. These datasets feature tabular data with a mix of categorical and continuous variables. Detailed statistical characteristics are presented in Table \ref{tab:statics}.
{
\setlength{\abovecaptionskip}{0pt}
\vspace{-0.5em}
\begin{table}[htbp]
\vspace{0.0em}
  \centering
  \caption{Statistics of the datasets in terms of number of examples, number of features, number of classes.}
  \small
  \setlength{\tabcolsep}{6pt} 
  \begin{tabular}{lccc} 
    \toprule[2pt]
    \textbf{Datasets} & \textbf{\# of examples} & \textbf{\# of features} & \textbf{\# of classes} \\
    \midrule[1pt]
    adult & 48842 & 14    & 2     \\
    arrhythmia & 452   & 279   & 16    \\
    crowdsourced& 10546 & 29 & 6 \\
    nsl-kdd & 148517 & 42    & 5     \\
    academic& 4424 & 36 & 3 \\
    accelerometer& 31991 & 8 & 2 \\
    metro& 1516948 & 15 & 2 \\
    diabetes& 101766 & 47 & 3 \\
    \bottomrule[2pt]
  \end{tabular}
  \vspace{-1.0em}
  \label{tab:statics}
\end{table}
\vspace{-0.1em}
}

\textbf{Settings. }
For comparison, we widely choose 10 models from diverse families, including Random Forest (RF) \cite{breiman2001random}, prevalent gradient boosting trees such as XGBoost \cite{chen2016xgboost}, LightGBM \cite{ke2017lightgbm}, and CatBoost \cite{prokhorenkova2018catboost}. We also compare with DANET \cite{chen2022danets}, a state-of-the-art neural network model for tabular modeling, as well as AutoGluon \cite{mueller2020faster}, which constructs crazy ensembles with thousands of base learners across model families with AutoML technique. Additionally, we evaluated the vanilla Deep Forest alongside improved variants csDF \cite{pang2018improving}, mdDF \cite{lyu2019refined}, and hiDF \cite{chen2021improving}.

Furthermore, due to the limited support for multi-output training in RF, which is essential for CMT module, we replaced RF with the SketchBoost \cite{iosipoi2022sketchboost} in AugDF, which is a fast implemention for multioutput GBDT. To isolate the effect, we introduced a variant named skDF, which simply replaces the RF in DF with SketchBoost.

For all experiments, same train-test splits are maintained, ensuring consistency in evaluation. We conduct each experiment five times with different random seeds and report the average performance along with the standard deviation. All DFs have a max layer of 15, and the number of individuals is set to 8 during the Augmentation Policy Schedule Learning.


For the sake of reproducibility, we have made the source code and parameter settings for this work publicly available at \url{https://github.com/dbsxfz/AugDF}.

\begin{table*}[htbp]
  \centering
  
  \vspace{-1.0em}
  \caption{Comparison of test accuracy of each models across datasets. The best accuracy is highlighted in \textbf{Bold} type. $\bullet$ indicates the second-best. The average rank is listed at the bottom.}
  \resizebox{\textwidth}{!}{
    \begin{tabular}{lllllllllllll}
    \toprule[2pt]
    \textbf{Datasets} & \multicolumn{1}{c}{\textbf{DANET}} & \multicolumn{1}{c}{\textbf{RF}} & \multicolumn{1}{c}{\textbf{LightGBM}} & \multicolumn{1}{c}{\textbf{XGboost}} & \multicolumn{1}{c}{\textbf{DF}} & \multicolumn{1}{c}{\textbf{mdDF}} & \multicolumn{1}{c}{\textbf{hiDF}} & \multicolumn{1}{c}{\textbf{skDF}} & \multicolumn{1}{c}{\textbf{csDF}} & \multicolumn{1}{c}{\textbf{Catboost}} & \multicolumn{1}{c}{\textbf{AutoGluon}} & \multicolumn{1}{c}{\textbf{AugDF}} \\


    \midrule[1pt]
    \textbf{adult} & 85.11±0.15 & 85.60±0.03 & 87.26±0.05 & 86.82±0.08 & 86.08±0.07 & 86.87±0.09 & 86.18±0.07 & 87.17±0.06 & 86.20±0.13 & 87.29±0.00 & 87.42±0.04$\bullet$ & \textbf{87.58±0.02} \\
    \textbf{arrhythmia} & 65.54±1.57 & 72.07±0.57 & 71.17±1.61 & 74.59±0.67 & 75.50±0.36 & 71.71±1.22 & 74.77±0.57 & 74.95±1.75 & 75.50±1.20 & 75.14±0.72 & 77.48±0.00$\bullet$ & \textbf{77.66±0.88} \\
    \textbf{crowd} & 61.33±1.80 & 64.27±0.44 & 65.27±0.80 & 62.53±1.07 & 62.87±0.62 & 65.47±0.62 & 63.13±0.62 & 63.53±1.75 & 63.67±0.87 & 66.27±0.98$\bullet$ & 65.73±0.49 & \textbf{67.20±0.58} \\
    \textbf{kdd} & 76.11±0.53 & 74.32±0.16 & 75.01±0.22 & 77.09±0.08 & 77.11±0.52 & 76.88±0.21 & 76.73±0.54 & 76.13±0.16 & 77.10±0.31 & 77.33±0.13 & 77.41±0.07$\bullet$ & \textbf{78.89±0.12} \\
    \textbf{academic} & 74.12±0.78 & 77.20±0.37 & 76.43±0.15 & 76.36±0.34 & 76.61±0.34 & 76.47±0.19 & 76.97±0.55 & 77.04±0.39 & 76.50±0.10 & 76.41±0.50 & 77.20±0.24$\bullet$ & \textbf{77.92±0.18} \\
    \textbf{accelero} & 98.36±0.08 & 98.52±0.02 & 98.49±0.02 & 98.45±0.05 & 98.54±0.02 & 98.54±0.03 & 98.55±0.02 & 98.51±0.06 & 98.54±0.04 & 98.55±0.05$\bullet$ & 98.54±0.01 & \textbf{98.57±0.01} \\
    \textbf{metro} & 98.23±0.37 & 98.73±0.01 & 98.51±0.04 & 98.74±0.05 & 98.41±0.29 & 98.42±0.01 & 98.68±0.14 & 98.56±0.21 & 98.62±0.13 & 98.78±0.02$\bullet$ & 98.60±0.04 & \textbf{99.15±0.14} \\
    \textbf{diabetes} & 58.11±0.14 & 58.59±0.05 & 59.31±0.05 & 59.20±0.09 & 58.64±0.09 & 58.95±0.06 & 58.73±0.05 & 59.28±0.10 & 58.75±0.13 & 59.10±0.07 & 59.54±0.05$\bullet$ & \textbf{59.70±0.07} \\

    \midrule[1pt]
    Avg.Rank& 11.75 & 7.94  & 7.75  & 7.75  & 7.31  & 7.06  & 6.88  & 6.63  & 6.38  & 4.125 & 3.44$\bullet$&\textbf{1.00}\\
    \bottomrule[2pt]
    \end{tabular}%
    }
    \vspace{-1.0em}
  \label{tab:com}%
\end{table*}%
\label{sec:majhead}
\vspace{-15pt}
\subsection{Test Accuracy on Benchmark Datasets}
\vspace{-5pt}
\label{sec:3.1}

Results in Table \ref{tab:com} demonstrate that AugDF consistently outperforms other models in terms of classification accuracy across all eight datasets. The performance gap between AugDF and the vanilla DF is particularly noteworthy. While skDF, which replaces the base learners, outperforms vanilla DF, it does not surpass csDF and Catboost due to overfitting and the challenge to determine the optimal output layer. AugDF demonstrates a robust performance uplift, irrespective of base learner replacement, achieving an average improvement of approximately 2\% over skDF on most datasets.
Interestingly, even the extensive AutoML ensemble, AutoGluon, fails to match AugDF's performance, underscoring the efficacy of AugDF's deep representation and the benefits derived from its learnable data augmentation policy schedule.


\vspace{-15pt}
\subsection{Comparison of Model Training and Inference Time}
\vspace{-5pt}
\label{sec:3.2}

In this section, we juxtapose the training and inference latencies of all considered models and employ scatter plots to visualize average performance metrics across the eight datasets. As depicted in Figure \ref{time}, it is evident that AugDF achieves optimal performance with a moderate computational cost. AutoGluon, while achieving suboptimal performance compared to AugDF, incurs over five times the training cost and exponentially higher inference latency. When compared to skDF, which utilizes the same base learners, AugDF's training cost is only about tenfold higher within an expansive search space of $10^{30}$. Considering that the search process is highly parallelizable, the time complexity may only slightly increase when executed on efficient parallel computing frameworks \cite{222605}.
It is worth noting that AugDF demonstrates superior inference efficiency compared to DF and its variants, primarily due to the shallower trees used by SketchBoost in AugDF.

{
\vspace{-5pt}
\begin{figure}[htbp]
\begin{minipage}[b]{1.0\linewidth}
  \centering
  \centerline{\includegraphics[width=\linewidth]{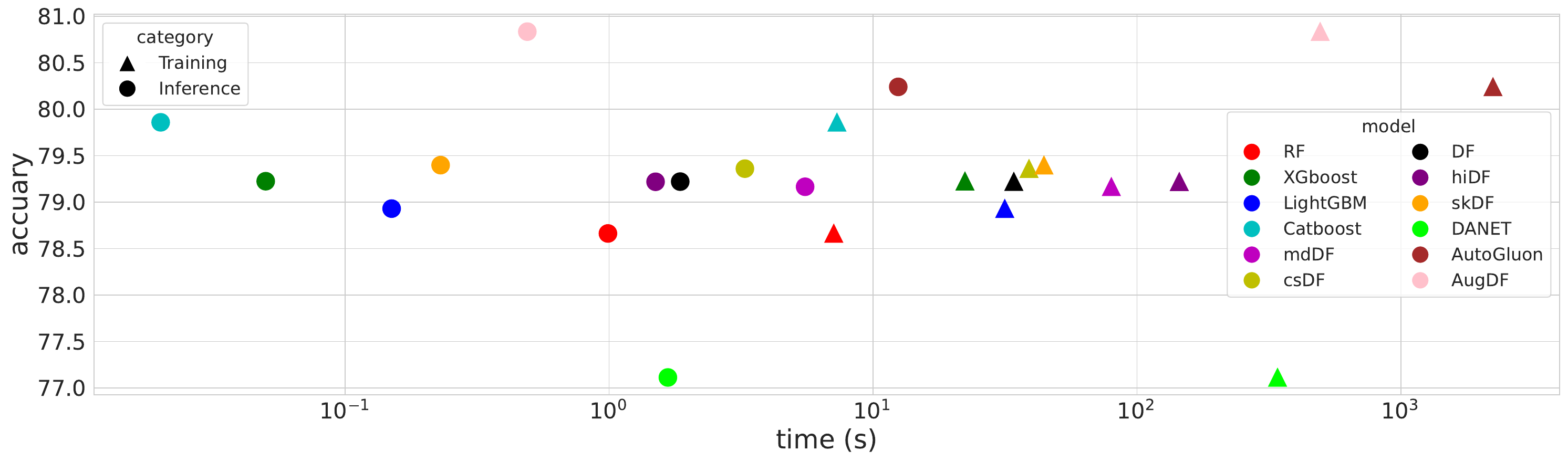}}
  \vspace{-5pt}
\end{minipage}
\caption{Scatter Plot of Model Accuracy and Latency (Logarithmic Scale). The values displayed represent averages across these eight datasets.}
\label{time}
\end{figure}

}

\subsection{Transferring Policy Schedule to Variants of DF}
\vspace{-5pt}
Tree-based models are inherently non-parametric, posing challenges for transfer learning \cite{wen2021challenges}. In this section, we directly apply the policy schedules discovered by AugDF to three variants of DF to further demonstrate the effectiveness and generalizability of our approach. As illustrated in Figure \ref{transfer}, the policy schedules are universally beneficial, with all DF variants exhibiting positive improvements across all datasets, albeit less so than the gains seen with AugDF over DF due to AugDF's exhaustive policy search. The transfer of augmentation policy schedules introduces a novel facet of transfer learning, which is especially advantageous for high-cost training models like hiDF. 
{
\vspace{-8pt}
\begin{figure}[htbp]
\begin{minipage}[b]{1.0\linewidth}
  \centering
  \centerline{\includegraphics[width=\linewidth]{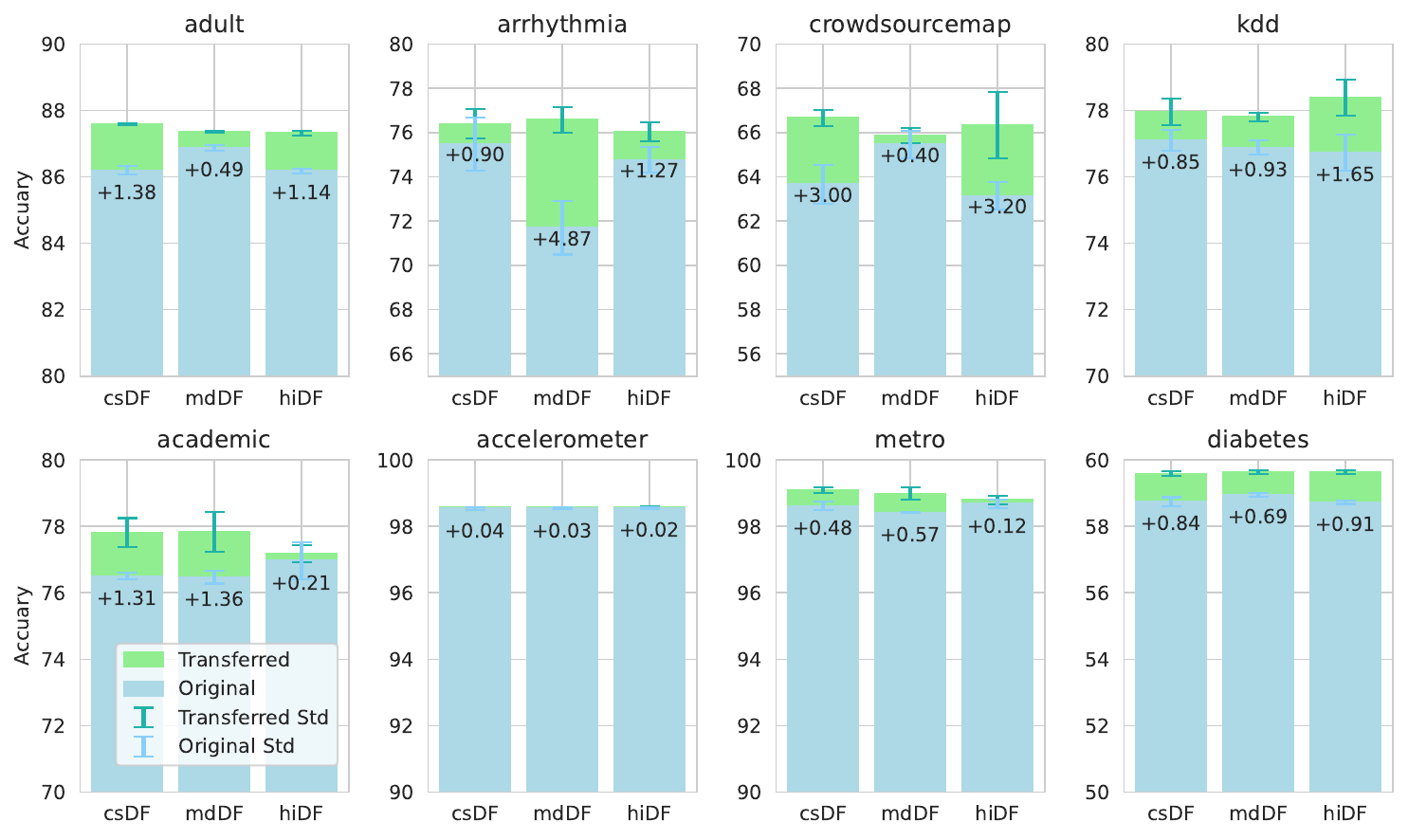}}
  \vspace{-5pt}
\end{minipage}
\caption{Comparative Analysis of Accuracy Improvement with Transferred Augmentation Policy Schedules. }
\label{transfer}
\end{figure}

}

{
\vspace{-2.0em}
\section{CONCLUSION}
\vspace{-0.8em}
}
In conclusion, this paper introduces an improved Deep Forest training approach specifically designed for tabular signal classification. Through the integration of CMT data augmentation technique, population-based augmentation policy schedule learning and checkpoint ensemble, we successfully mitigate overfitting and elevate model performance. These methodological advancements enable us to achieve SOTA results across a variety of benchmarks. Notably, the learned augmentation policy schedules are not only effective but also transferable, allowing them to be applied to variants of Deep Forest. This demonstrates its potential for broader impact in the field of non-differentiable deep learning.



\vfill\pagebreak
\renewcommand{\bibfont}{\footnotesize} 
\bibliographystyle{IEEEbib}
\setlength{\bibsep}{2pt} 
\bibliography{refs}

\end{document}